\documentclass[conference, letterpaper]{IEEEtran}

\usepackage{float}

\usepackage[pdftex]{graphicx}
\DeclareGraphicsExtensions{.pdf,.jpeg,.png}

\ifCLASSINFOpdf
\else
\fi
\ifCLASSINFOpdf
   \usepackage[pdftex]{graphicx}
\else
\fi

\usepackage[cmex10]{amsmath}
\usepackage{color}
\usepackage{fancyhdr}
\usepackage[caption=false,font=footnotesize]{subfig}

\renewcommand{\thispagestyle}[2]{}

\fancypagestyle{plain}{
        \fancyhead{}
        \fancyhead[C]{first page center header}
        \fancyfoot{}
        \fancyfoot[C]{first page center footer}
}
\begin{document}

\title{CIFAR-10: KNN-based Ensemble of Classifiers}

% \author{\IEEEauthorblockN{Yehya Abouelnaga}
% \IEEEauthorblockA{School of Sciences and Engineering\\
% The American University in Cairo\\
% New Cairo 11835, Egypt\\
% devyhia@aucegypt.edu}
% \and
% \IEEEauthorblockN{Ola S. Ali}
% \IEEEauthorblockA{School of Sciences and Engineering\\
% The American University in Cairo\\
% New Cairo 11835, Egypt\\
% 0la132@aucegypt.edu}
% \and
% \IEEEauthorblockN{Hager Rady}
% \IEEEauthorblockA{School of Sciences and Engineering\\
% The American University in Cairo\\
% New Cairo 11835, Egypt\\
% hagerradi@aucegypt.edu}}


\author{\IEEEauthorblockN{Yehya Abouelnaga, Ola S. Ali, Hager Rady, and Mohamed Moustafa}
\IEEEauthorblockA{Department of Computer Science and Engineering, School of Sciences and Engineering\\
The American University in Cairo, New Cairo 11835, Egypt\\
\{ devyhia , olasalem1 , hagerradi , m.moustafa \}@aucegypt.edu}
}


\maketitle

\begin{abstract}
%\boldmath
In this paper, we study the performance of different classififers on the CIFAR-10 dataset, and build an ensemble of classifiers to reach a better performance.
We show that, on CIFAR-10, K-Nearest Neighbors (KNN) and Convolutional Neural Network (CNN), on some classes, are mutually exclusive, thus yield in higher accuracy when combined.
We reduce KNN overfitting using Principal Component Analysis (PCA), and ensemble it with a CNN to increase its accuracy.
Our approach improves our best CNN model from 93.33\% to 94.03\%.

\end{abstract}


\begin{IEEEkeywords}
Ensemble of Classifiers; K-Nearest Neighbors; Convolutional Neural Networks; Principal Component Analysis
\end{IEEEkeywords}

% \fi
\IEEEpeerreviewmaketitle

\section{Introduction}
CIFAR-10 is a multi-class dataset consisting of 60,000 $32 \times 32$ colour images in 10 classes, with 6,000 images per class. There are 50,000 training images and 10,000 test images \cite{Krizhevsky2009}.
In this paper, we explore different learning classifiers for the image-based multi-class problem.
We will begin with training simple classifiers (like Logisitc Regression and Bayesian), and incremently move towards more complex alternatives: Support Vector Machines, Decision Trees, Random Forrests, Gradient Boosting and K-Nearest Neighbours.
Eventually, we will train a Deep Convolutional Neural Network (CNN).
We will also explore various feature engineering appraoches like Principal Component Analysis (PCA).
In an attempt to improve the state-of-the-art accuracy, we will devise an ensemble of classifiers.
We will compare the performance of the previously mentioned classifiers, feature extractors, and their effect on the final ensemble.
\par
CIFAR-10 presents a challenging classification problem.
$32 \times 32$ images don't contain enough information for most classifiers to draw clear decision boundaries.
A clear example of this is the confusion between ``cat" and ``dog" classes.
The images objects are different in scale, rotation, position, and background.
Some of the images are very unclear and hard to classify (even for human beings \cite{Karpathy2011}).

\section{Literature Review}
Neural networks are widely used in solving image recognition problems.
There is a wide variety of architectures that serve different purposes.
Some publications aim at simple architectures to achieve decent results.
\cite{mcdonnell2015enhanced} presents a shallow neural network that is, unlike multi-layered (i.e. deep) architectures, fast to train and more suitable for realtime applications.
Their network achieved 75.86\% test accuracy.

\cite{chan2015pcanet}, \cite{coates2012learning} and \cite{dosovitskiy2014discriminative} optimize their networks to learn better representation of features.
\cite{chan2015pcanet} presents a simple architecture (PCANet) where layers of PCA is used to learn features (rather than using convolutional layers).
PCANet achieves 78.67\% test accuracy.
\cite{coates2012learning} suggests using K-means (unsupervised learning) to learn better feature representations to transform the image space into a linearly separable feature space to be used with a standard linear classification algorithm (e.g. SVM).
\cite{dosovitskiy2014discriminative} aims at learning features by training a convolutional neural network using only unlabelled data.

% \cite{hinton2012improving}, \cite{goodfellow2013maxout}, \cite{springenberg2013improving} and \cite{chang2015batch} try to reduce overfitting by model averaging techniques.
% \cite{hinton2012improving} introduces \textit{dropout} (where “overfitting” is greatly reduced by randomly omitting half of the feature detectors on each training case).

\cite{malinowski2013learning}, \cite{zeiler2013stochastic}, \cite{graham2014fractional} and \cite{lee2016generalizing} experiment with network pooling.
\cite{malinowski2013learning}, \cite{zeiler2013stochastic} and \cite{graham2014fractional} suggest regulazing existing pooling functions.
\cite{malinowski2013learning} proposes a flexible parameterization of the spatial pooling step and learn the pooling regions together with the classifier.
\cite{zeiler2013stochastic} replaces the conventional deterministic pooling operations with a stochastic procedure (randomly picking the activation within each pooling region).
\cite{graham2014fractional} has formulated a fractional (stochastic) version of maxpooling (where non-integer multiplicative factors are allowed).
This helps reduce overfitting, and achives state-of-the-art accuracy (with 96.53\% test accuracy).
\cite{lee2016generalizing} learns new pooling functions by combination of max and average pooling functions, or tree-structured fusion of pooling filters.

\cite{xu2015empirical}, \cite{agostinelli2014learning}, \cite{clevert2015fast}, \cite{goodfellow2013maxout}, \cite{springenberg2013improving} and \cite{tang2013deep} optimize network activation functions.
\cite{xu2015empirical} presents a new randomized leaky rectified linear units (RReLU).
\cite{agostinelli2014learning} designed a novel form of piecewise linear activation function that is learned independently for each neuron using gradient descent (and outperforms rectified linear units).
\cite{clevert2015fast} introduce an exponential linear unit (ELU) which speeds up learning in deep neural networks and leads to higher classification accuracies.
\cite{goodfellow2013maxout} defines a simple new model called \textit{maxout} to improve the accuracy of \textit{dropout} (see \cite{hinton2012improving}) fast approximate model averaging technique.
\cite{springenberg2013improving} present a probabilistic variant (\textit{probout}) of the recently introduced \textit{maxout} unit to improve its invariance properties.
\cite{tang2013deep} shows that replacing the softmax layer with a linear support vector machine (SVM) consistently improves accuracies.

\cite{lin2013network} introduces a novel deep network structure called ``Network In Network" (NIN) to enhance model discriminability for local patches within the receptive field.
\cite{chang2015batch} combines Network in Network architecture (NIN) (see \cite{lin2013network}) and the maxout units (see \cite{goodfellow2013maxout}) to enhance model discriminability and facilitate the process of information abstraction within the receptive field.

\cite{visin2015renet} proposes a deep neural network architecture for object recognition based on recurrent neural networks (ReNet).
The proposed network replaces the ubiquitous convolution+pooling layer of the deep convolutional neural network with four recurrent neural networks that sweep horizontally and vertically in both directions across the image.
\cite{liang2015recurrent} proposes a recurrent CNN (RCNN) for object recognition by incorporating recurrent connections into each convolutional layer to enhance the ability of the model to integrate the context information (which is important for object recognition).

\cite{zagoruyko2016wide} introduces a novel architecture that decreases depth and increases width of residual networks.
Wide residual networks (WRNs) are superior to their deep residuals couterparts.
\cite{mishkin2015all} proposes a simple method for weight initialization for deep net learning, called Layer-sequential unit-variance (LSUV) initialization to improve test accuracies.

The previous publications aim at improving neural networks as general learning tools.
They research new activation functions, pooling functions, weight initialization methods, learning algorithms, and variations of existing models.
Ensemble of classifiers are proven to achieve better results in the literature (see \cite{Opitz1999,Rahman2014,Dietterich}).
Despite their robustness, variations of CNN-based ensembles are not thoroughly researched and analyzed.
In an attempt to explore more variations of ensembles, we analyze the impact of ensembling a simple learning tool like K-Nearest Neighbours (KNN) with Convolutional Neural Networks (CNN).

% Simplification
% Feature Learning
% Pooling

% Overfitting Imrpovement (Dropout)
% \cite{hinton2012improving}
% \cite{goodfellow2013maxout} (Maxout .. builds on Dropout)
% \cite{springenberg2013improving} (Probabilistic Maxout units)
% \cite{chang2015batch} (Maxout Network In Network)

% Optimizing Activation Functions
% \cite{xu2015empirical}
% \cite{agostinelli2014learning}
% \cite{clevert2015fast} (New activation unit)
% \cite{tang2013deep}

% Recurrent Neural Network Alternative
% \cite{visin2015renet}
% \cite{liang2015recurrent} (Recurrent Deep Neural Network)

% Non Neural Network Approaches
% \cite{vinyals2012learning}

% Wide \& Deep DNN
% \cite{ciregan2012multi}

% ImageNet training (and CIFAR-10 test)
% \cite{krizhevsky2012imagenet}

% Network In Network (New NN model)
% \cite{lin2013network}
% \cite{chang2015batch} (Maxout Network In Network) (and Dropout)

% Weight Initialization
% \cite{mishkin2015all} (All you need is a good init)

\section{Proposed Method}
We propose an ensemble-based approach to improve the CIFAR-10 test accuracy.
We experiment with possible learning models, and try to find the best combination of classifiers to reduce classifier confusion and improve accuracy.
We found that K-Nearest Neighbours (KNN) consistently improves the accuracy of Convolutional Neural Networks (CNN).

\begin{figure}[H]
\centering
\includegraphics[width=3in]{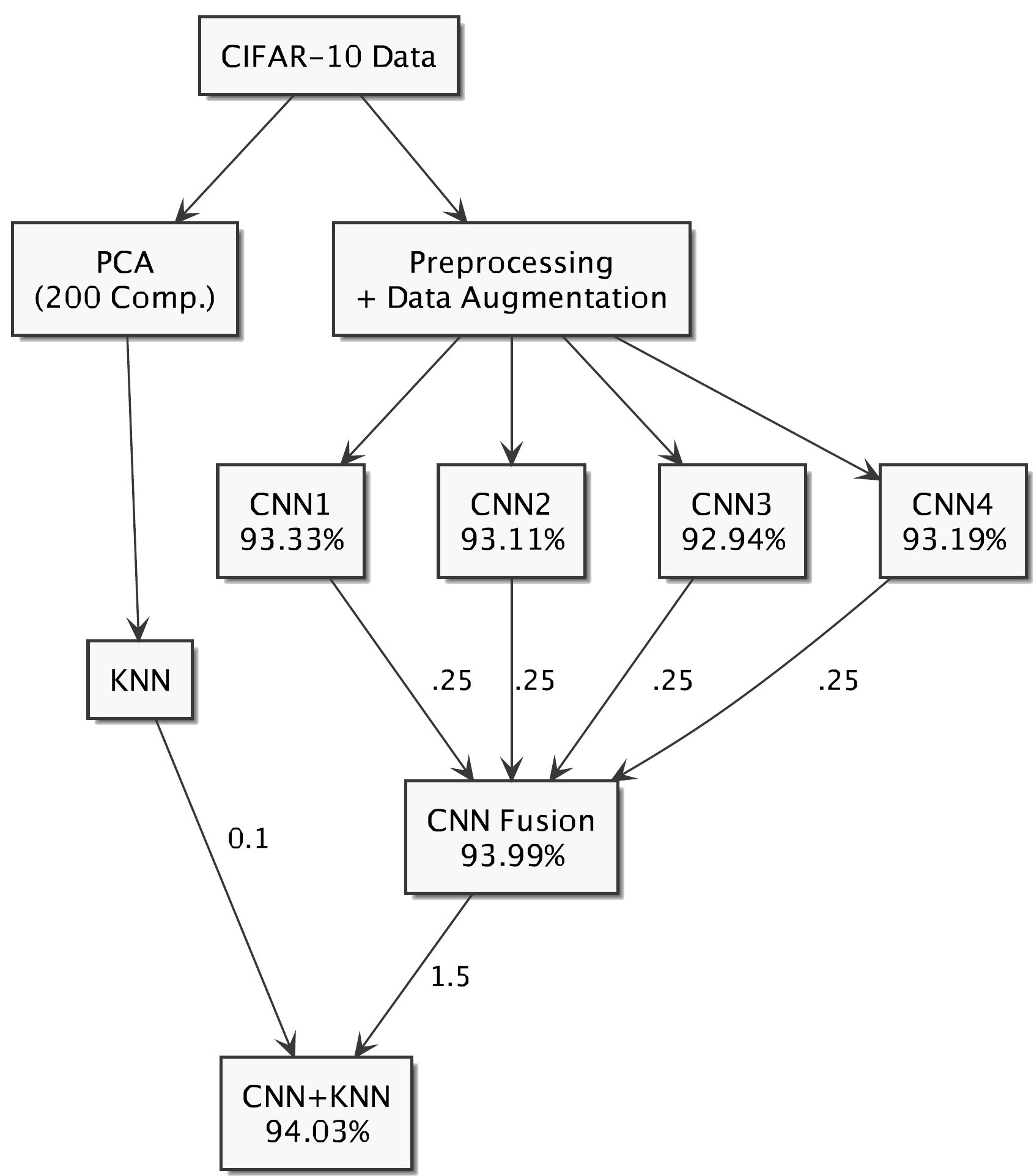}
\caption{Architecture of ensembling 4 ConvNets and KNN. PCA improves the accuracy of KNN as it reduces overfitting. Ensembling the 4 ConvNets improves their accuracy to 93.99\%. KNN improves the accuracy of the ensemble to 94.03\%.}
\end{figure}

\subsection{Principal Component Analysis (PCA)}
 PCA is a dimensionality reduction tool used to remove noise from images.
 This behaviour is very clear when using K-Nearest Neighbours (KNN) for classification.
 The lower the number of components, the more accurate the distance function performs.
 Throughout this paper, PCA will provide performance gains in almost all classifiers.

\begin{figure}[!t]
\centering
\includegraphics[width=2.5in]{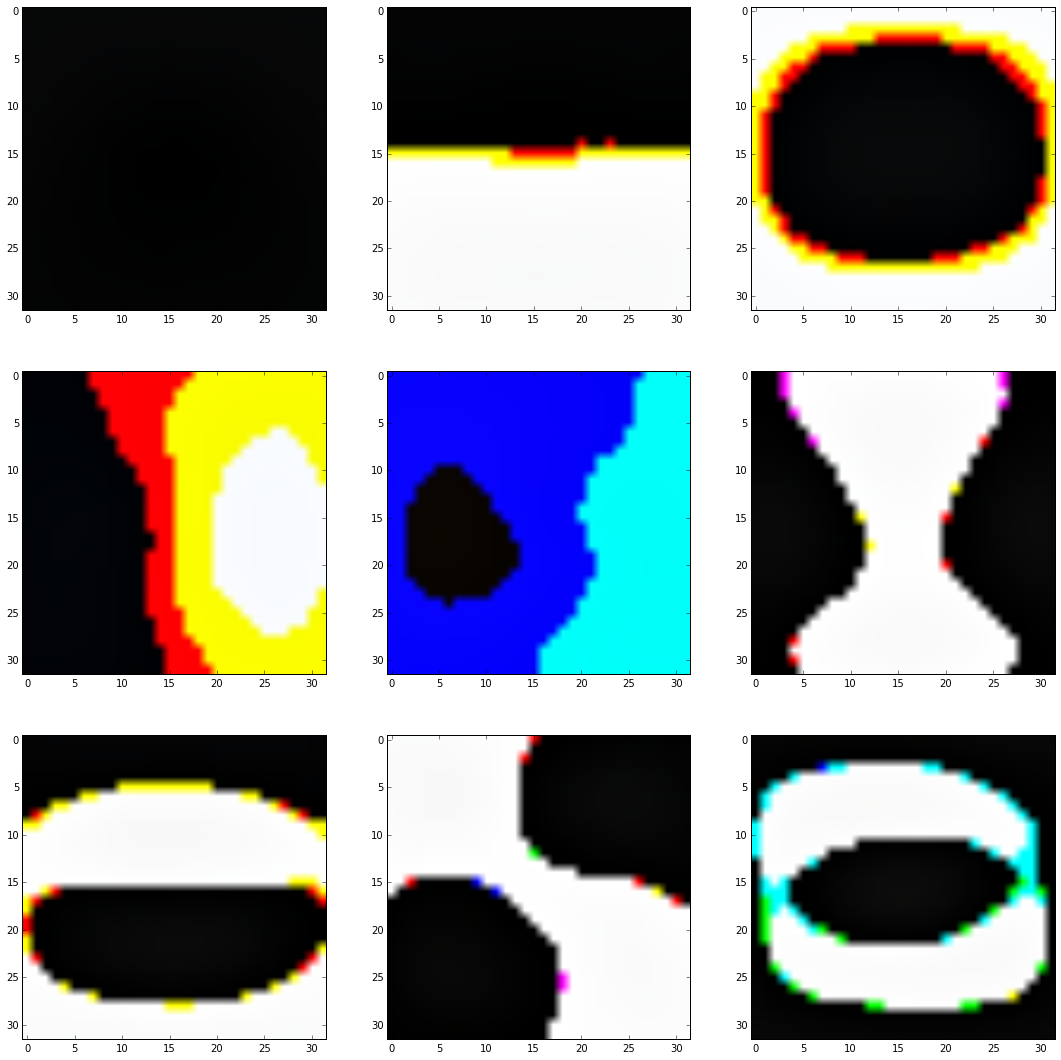}
\caption{The first 9 components in the PCA (of 200 components) preserve 65.5\% of the data.}
\end{figure}

\begin{figure}[!t]
\centering
\includegraphics[width=2.5in]{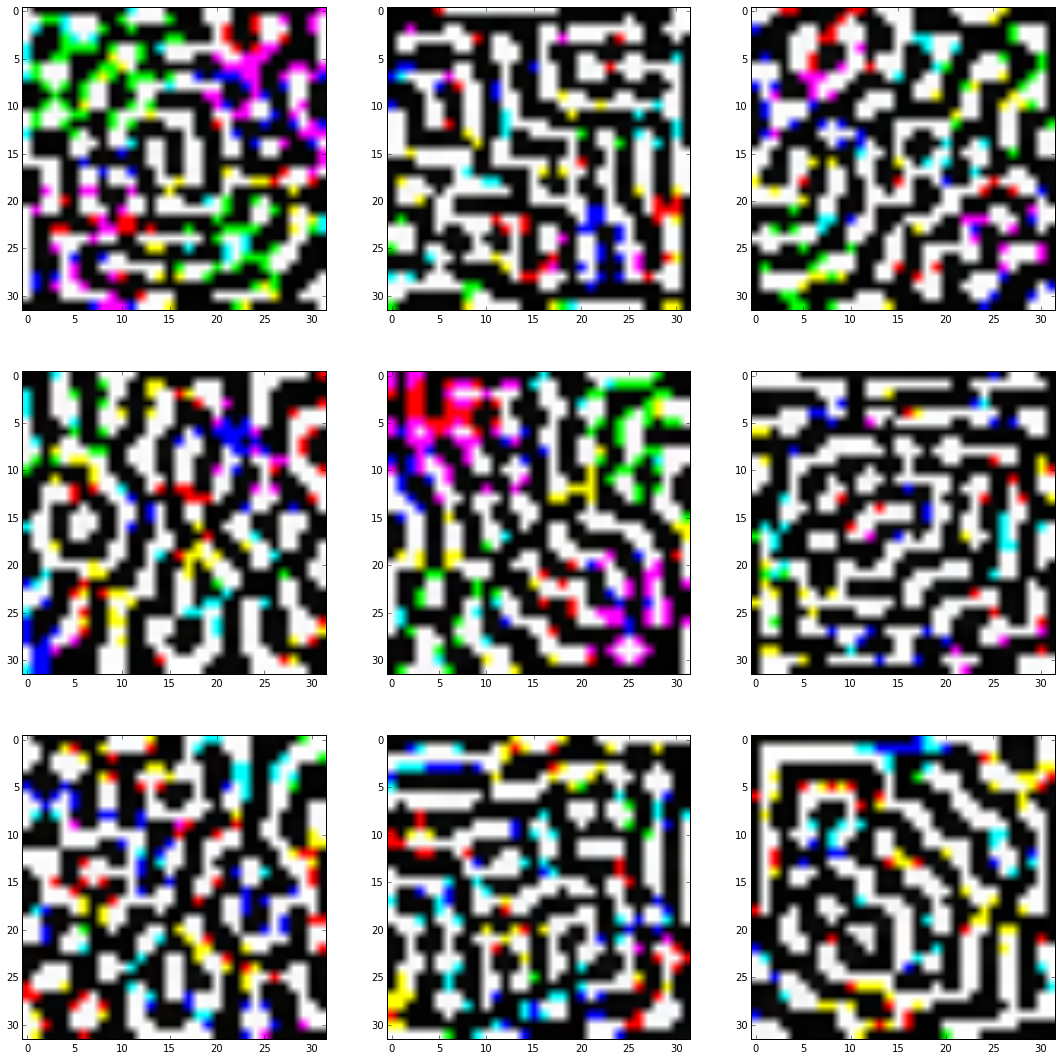}
\caption{The last 9 components in the PCA (of 200 components) preserve 0.28\% of the data.}
\end{figure}

\subsection{K-Nearest Neighbours}
K-Nearest Neighbours (KNN) is a simple non-parametric classification method that depends on the between-sample geometric distance.
It finds the best $k$ means, $M_1, ..., M_k$, of the training sample representing the entire sample, and classifies every new point $X_i$ based on the closest distance, $D$, to the means.
\[
C(X_i) = \underset{k}{\text{argmin}} \,\, D(X_i, M_k)
\]
The simplicity of the KNN model rectifies major confusion between similar classes (i.e. cats, dogs and horses).

\subsection{Convolutional Neural Network (CNN)}
We used the best available model (see \cite{Nagadomi2014}).
Models from other publications with higher accuracies were either not publicly available or not training properly.
The architecture of our deep neural network consists of 8 convolutional layers in addition to 3 linear layers (see Table \ref{conv:arch}).
It achieves, on average, a test accuracy of 93.13\%. We trained the model 4 times with different initial seeds and averaged them to produce 93.99\% test accuracy.
\par
Data augmentation has been used to artificially enlarge the size of the dataset and reduce the effect of over fitting.
We used cropping, horizontal reflection (similar to \cite{krizhevsky2012imagenet}) and scaling.
All three methods allow the transformed image to be generated from the original image with little computation.
As for preprocessing, we used Global Contrast Normalization (GCN) and ZCA whitening.

\begin{figure}[!t]
  \label{cnn_arch}
\centering
\includegraphics[width=3.5in]{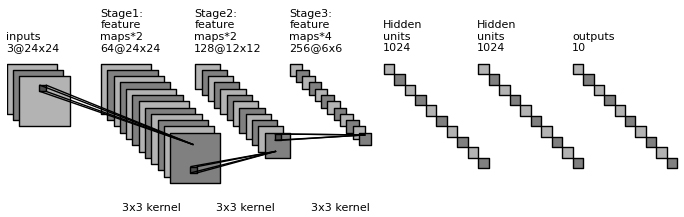}
\caption{Convolutional Neural Network Architecture.}
\end{figure}

\begin{table}[!t]
\centering
\caption{Convolutional Neural Network Architecture}
\label{conv:arch}
\begin{tabular}{|c|}
\hline
Input                                                                                               \\ \hline
\begin{tabular}[c]{@{}c@{}}2 x (Conv. + ReLU)\\ kernel: 3x3, channel: 64, padding: 1\end{tabular}   \\ \hline
\begin{tabular}[c]{@{}c@{}}Max Pooling (kernel: 2x2, stride: 2)\\ Dropout (rate: 0.25)\end{tabular} \\ \hline
\begin{tabular}[c]{@{}c@{}}2 x (Conv. + ReLU)\\ kernel: 3x3, channel: 128, padding: 1\end{tabular}  \\ \hline
\begin{tabular}[c]{@{}c@{}}Max Pooling (kernel: 2x2, stride: 2)\\ Dropout (rate: 0.25)\end{tabular} \\ \hline
\begin{tabular}[c]{@{}c@{}}4 x (Conv. + ReLU)\\ kernel: 3x3, channel: 256, padding: 1\end{tabular}  \\ \hline
\begin{tabular}[c]{@{}c@{}}Max Pooling (kernel: 2x2, stride: 2)\\ Dropout (rate: 0.25)\end{tabular} \\ \hline
Linear (channel: 1024) + ReLU                                                                       \\ \hline
Dropout (rate: 0.5)                                                                                 \\ \hline
Linear (channel: 1024) + ReLU                                                                       \\ \hline
Dropout (rate: 0.5)                                                                                 \\ \hline
Linear (channel: 10)                                                                                \\ \hline
Softmax                                                                                             \\ \hline
\end{tabular}
\end{table}

\subsection{Ensemble Weight Estimation}
\cite{Opitz1999,Kim2011,Rahman2014,Dietterich} propose multiple approaches for ensemble parameter estimation in a weighted voting system.
However, we opted out for a simple exhaustive search.
Assume, $C_1, C_2, ..., C_n$, are experts.
We define a sequence of possible weights, $W_{n+1} = W_n + S$ and $W_0 = 0$, where $S$ is a step between two consecutive weights.
Let $R = \{W_0, W_1, ..., W_k\}$, where $k$ is the number of possible weights.
\[
E(C_1, C_2) = \underset{w_i, w_j \in R}{\text{argmax}} \,\, w_i \times C_1 + w_j \times C_2
\]
We estimate all parameters in a chain rule style (i.e. $E(E(C_1, C_2), C_3)$).

\section{Experiments}

\subsection{Feature Extractors}
We experimented with Principal Component Analysis (PCA), Scale-Invariant Feature Transform (SIFT),  Speeded Up Robust Features (SURF), Oriented Fast and Rotated Brief (ORB), and Histogram of Oriented Gradients (HOG).
All, except PCA, overfitted the model.
They increased the training accuracy but lowered the test accuracy.
For most classifiers (i.e. Logistic Regression, Decision Trees, Random Forests, and Gradient Boosting), we found that 200 PCA components (out of 3,072) reduce overfitting and increase test accuracy.
For K-Nearest Neighbours, the best test accuracy was acheived by a lower number of components (30 components).

\subsection{Results Analysis}
\par
\begin{figure}[!t]
\centering
\includegraphics[width=3.5in]{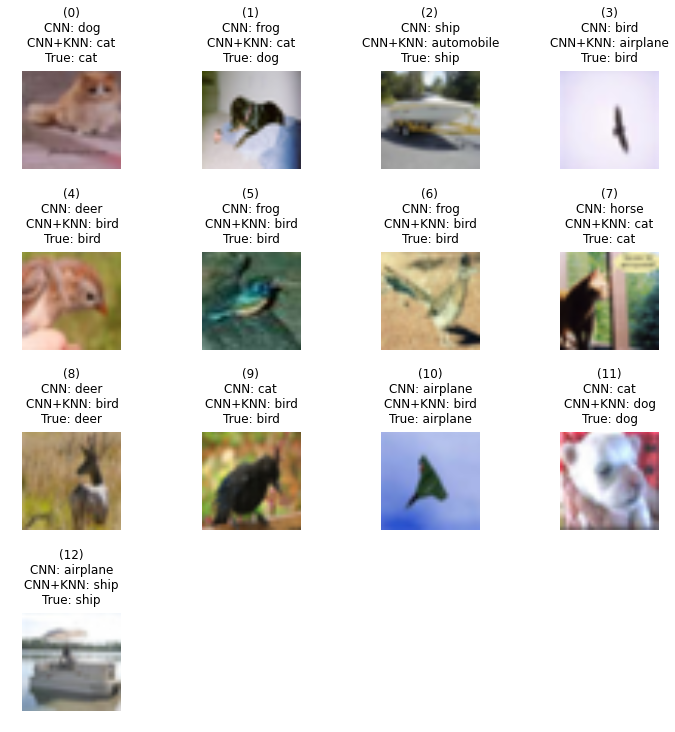}
\caption{Effect of K-Nearest Neighbours (KNN) on our Convolutional Neural Network (CNN) model.}
\label{cnn_knn_images}
\end{figure}

In Fig. \ref{cnn_knn_images}, we study the difference between the CNN and KNN.
In images (0, 4, 5, 6, 7, 9, 11, 12), we find that KNN's vote in the classification decision improves CNN's insensitivity to shape.
It improved the confusion between cat, dog, and horse classes.
In image (1), it also predicted a dog as a cat (which is closer than a frog).
In image (2), it didn't generalize very well.
CNN voted for a ship. CNN+KNN voted for an automobile.
However, the image contains a ship on top of a vehicle (with two tires).
The shape sensitivity of the KNN vote helped reduce confusion among cat, horse, dog, and bird classes.
However, it confused the CNN on images (3, 8, 10), where airplane was confused for a bird (and vice versa), and a deer confused for a bird (due to its posture).

% \begin{figure}[!t]
% \label{fig:confusion}
% \centering
% \includegraphics[width=3.5in]{confusion}
% \caption{Convolutional Neural Network Confusion Matrix. A substantial error result from confusion between cats and dogs. 88 cat images are classified as dogs. 66 dog images are classified as cats.}
% \end{figure}

\begin{table}[!t]
\caption{Experimental Results}
\centering
\begin{tabular}{|c|c|}
\hline
Classifier & Accuracy (\%)\\
\hline
Log. Reg. + 3,072 Features & 37.5\\
\hline
Log. Reg. + 50 PCA Comp. & 37.69\\
\hline
Log. Reg. + 100 PCA Comp. & 	40.13\\
\hline
Log. Reg. + 150 PCA Comp. & 	40.18\\
\hline
\textbf{Log. Reg. + 200 PCA Comp.} & \textbf{41.04}\\
\hline
Log. Reg. + 225 PCA Comp. & 40.56\\
\hline
Log. Reg. + 250 PCA Comp. & 40.87\\
\hline \hline
KNN + 3,072 Features & 33.86 \\
\hline
KNN + 200 PCA Comp.   &  36.54 \\
\hline
KNN + 75 PCA Comp.  &  39.77 \\
\hline
KNN + 50 PCA Comp.  &  40.12 \\
\hline
KNN + 40 PCA Comp.  &  40.93 \\
\hline
\textbf{KNN + 30 PCA Comp.}  &  \textbf{41.78} \\
\hline
KNN + 25 PCA Comp.  &  41.57 \\
\hline
KNN + 15 PCA Comp.  &  38.75 \\
\hline
KNN + 10 PCA Comp.   &  34.93 \\
\hline
\hline
RFC/512   &  49.26 \\
\hline
RFC/1024  &  48.97 \\
\hline
RFC/512 + 200 PCA Comp.  &  48.59 \\
\hline
\textbf{RFC/1024 + 200 PCA Comp.} &  \textbf{49.52} \\
\hline
\hline
\textbf{GRB + 3,072 Features} & \textbf{47.78}\\
\hline
\hline
\textbf{SVM + 3,072 Features} & \textbf{49.88}\\
\hline
\hline
CNN1 + Data Augm. & 93.33\\
\hline
CNN2 + Data Augm. & 93.11\\
\hline
CNN3 + Data Augm. & 92.94\\
\hline
CNN4 + Data Augm. & 93.19\\
\hline
CNN Fusion & 93.99\\
\hline
\textbf{CNN Fusion + KNN} & \textbf{94.03}\\
\hline
\end{tabular}
\end{table}

\begin{table}[!t]
\centering
\caption{Ensembling Results}
% \label{my-label}
\begin{tabular}{|l|l|l|l|l|}
\hline
           & Base Line & KNN            & GRB            & RFC            \\ \hline
CNN1       & 93.33     & \textbf{93.46} & 93.38          & 93.40          \\ \hline
CNN2       & 93.11     & 93.15          & 93.15          & \textbf{93.16} \\ \hline
CNN3       & 92.94     & 92.97          & \textbf{92.98} & \textbf{92.98} \\ \hline
CNN4       & 93.19     & \textbf{93.25} & 93.19          & 93.21          \\ \hline
CNN Fusion & 93.99     & \textbf{94.03} & 94.00          & 93.99          \\ \hline
\end{tabular}
\end{table}

\section{Conclusion}
K-Nearest Neighbors (KNN) is a simple classification algorithm based on geometric distance.
On CIFAR-10 dataset, we showed that it stabilizes the decision of other Convolutional Neural Networks (CNN) due to its shape sensitivity.
We showed that an ensemble of CNNs and KNN improves the accuracy of the model.
More research is to be done on the effect of KNN on CNN.
A KNN needs to be checked against other datasets to generalize a statement on the effect of KNN.
Other weight estimation methods should be evaluated and compared to the simple exhaustive search we presented in this paper.

% The CIFAR-10 dataset consists of 60,000 $32 \times 32$ colour images in 10 classes, with 6,000 images per class. There are 50,000 training images and 10,000 test images \cite{Krizhevsky2009}. In this paper, we explore different learning classifiers for the image-based multi-class problem. We will begin with training simple classifiers (like Logisitc Regression and Bayesian), and incremently move towards more complex alternatives: Support Vector Machines, Decision Trees, Random Forrests, Gradient Boosting and K-Nearest Neighbours. Eventually, we will train a Deep Convolutional Neural Network (CNN).

\section*{Acknowledgment}
% \cite{jia2014caffe}
The authors relied on the implementation of Scikit Learn Python Library \cite{Pedregosa2011} and Torch in most of the experiments carried out in this paper.



% The CIFAR-10 dataset consists of 60,000 $32 \times 32$ colour images in 10 classes, with 6,000 images per class. There are 50,000 training images and 10,000 test images \cite{Krizhevsky2009}. In this paper, we explore different learning classifiers for the image-based multi-class problem. We will begin with training simple classifiers (like Logisitc Regression and Bayesian), and incremently move towards more complex alternatives: Support Vector Machines, Decision Trees, Random Forrests, Gradient Boosting and K-Nearest Neighbours. Eventually, we will train a Deep Convolutional Neural Network (CNN).



%\bibliographystyle{IEEEtran}
\bibliographystyle{IEEEtran}
\bibliography{Computing_2017_Paper.bib}
% \end{thebibliography}

\end{document}